\documentclass[letterpaper, 10 pt, journal, twoside]{IEEEtran}
\usepackage{enumitem}
\usepackage{wrapfig}
\usepackage{epsfig,xspace,layout}
\usepackage{color}
\usepackage{amsfonts}
\usepackage{times}
\usepackage{amssymb}
\usepackage{hyperref}
\usepackage{amsmath,bm}
\usepackage{rotating}
\usepackage{amsthm}
\usepackage{graphicx}
\usepackage{epsfig}
\usepackage{mathrsfs}
\usepackage{caption}
\usepackage{makecell}
\usepackage{colortbl}
\usepackage{mathtools}
\usepackage{makeidx}
\usepackage{multirow} 
\usepackage{dblfloatfix}
\usepackage{threeparttable}
\usepackage{dsfont}
\usepackage[font={small}]{caption}
\usepackage{algorithm}
\usepackage{algorithmic}
\usepackage{tikz}
\usepackage{siunitx}
\usepackage{multicol,lipsum}
\usepackage{dsfont}
\usepackage{amssymb}
\usepackage{url}
\usepackage{longtable}
\usepackage{booktabs}
\usepackage{lineno}
\usepackage{hhline}
\usepackage[Symbol]{upgreek}

\usepackage{colortbl}
\usepackage{amsmath,amsfonts}
\usepackage{algorithmic}
\usepackage{array}
\usepackage[caption=false,font=normalsize,labelfont=sf,textfont=sf]{subfig}
\usepackage{textcomp}
\usepackage{url}
\usepackage{verbatim}
\usepackage{graphicx}
\def\BibTeX{{\rm B\kern-.05em{\sc i\kern-.025em b}\kern-.08em
    T\kern-.1667em\lower.7ex\hbox{E}\kern-.125emX}}
\usepackage{balance}

\usepackage{graphicx}
\usepackage{caption}
\usepackage{subcaption}

\begin{document}
\title{ \Large \bf Learning Whole-Body Loco-Manipulation for Omni-Directional Task Space Pose Tracking with a Wheeled-Quadrupedal-Manipulator}
\author{Kaiwen Jiang$^{1}$, Zhen Fu$^{1}$, Junde Guo$^1$, Wei Zhang$^{1,3}$ and Hua Chen$^{2,3}$
\thanks{Manuscript received: August, 12, 2024; Revised November, 2, 2024; Accepted November, 25, 2024. This paper was recommended for publication by Editor Abderrahmane Kheddar upon evaluation of the Associate Editor and Reviewers' comments. \emph{(Kaiwen Jiang and Zhen Fu contributed equally to this work)(Corresponding author: Hua Chen)} }
\thanks{$^1$ School of System Design and Intelligent Manufacturing (SDIM), Southern University of Science and Technology, Shenzhen, China, 518055. Email: {\tt\footnotesize boocoly@gmail.com}, {\tt\footnotesize \{12131084, 
12332645\}@mail.sustech.edu.cn}, {\tt\footnotesize zhangw3@sustech.edu.cn}. 
}
\thanks{$^2$ Zhejiang University-University of Illinois Urbana-Champaign Institute (ZJUI), Haining, China, 314400. Email: {\tt\footnotesize huachen@intl.zju.edu.cn
}}
\thanks{$^3$ LimX Dynamics, Shenzhen, China, 518055}
\thanks{Digital Object Identifier (DOI): see top of this page.}
}

\markboth{IEEE Robotics and Automation Letters. Preprint Version. Accepted November, 2024}
{Jiang \MakeLowercase{\textit{et al.}}: Whole-Body Loco-Manipulation with a Wheeled-Quadrupedal-Manipulator} 

\maketitle
\begin{abstract}
In this paper, we study the whole-body loco-manipulation problem using reinforcement learning (RL). Specifically, we focus on the problem of how to coordinate the floating base and the robotic arm of a wheeled-quadrupedal manipulator robot to achieve direct six-dimensional (6D) end-effector (EE) pose tracking in task space. Different from conventional whole-body loco-manipulation problems that track both floating-base and end-effector commands, the direct EE pose tracking problem requires inherent balance among redundant degrees of freedom in the whole-body motion. We leverage RL to solve this challenging problem. To address the associated difficulties, we develop a novel reward fusion module (RFM) that systematically integrates reward terms corresponding to different tasks in a nonlinear manner. In such a way, the inherent multi-stage and hierarchical feature of the loco-manipulation problem can be carefully accommodated. By combining the proposed RFM with the a teacher-student RL training paradigm, we present a complete RL scheme to achieve 6D EE pose tracking for the wheeled-quadruped manipulator robot. Extensive simulation and hardware experiments demonstrate the significance of the RFM. In particular, we enable smooth and precise tracking performance, achieving state-of-the-art tracking position error of less than 5 cm, and rotation error of less than 0.1 rad. Please refer to \href{https://clearlab-sustech.github.io/RFM_loco_mani/}{this website} for more experimental videos.
\end{abstract}

\section{Introduction}
\IEEEPARstart{T}{hanks} to the exceptional versatility of accomplishing long-horizon multistage loco-manipulation tasks, control of mobile manipulator robotic systems has recently attracted substantial attention in the literature. Especially, reinforcement learning techniques have been shown to be promising in solving such complex problems. In this paper, we focus on the problem of how to coordinate the floating base and the robotic arm of a wheeled-quadrupedal manipulator robot to achieve direct six-dimensional (6D) end-effector (EE) pose tracking in task space. Different from conventional whole-body loco-manipulation problems that track both floating-base and end-effector commands, the direct EE pose tracking problem requires inherent balance among redundant degrees of freedom in the whole-body motion. 

Existing works such as~\cite{Fu2023a,Wang2024,Portela2024} have demonstrated that whole body RL policies are effective in controlling the three-dimensional position of the EE. However, these controllers often become highly customized for specific tasks and struggle with controlling the EE's orientation, restricting their generalization to a broader range of tasks or environments. To accomplish more versatile tasks, full 6D pose tracking is crucial. Tasks such as wiping the underside of a table, picking up a cup with a specific opening orientation, and straightening up a fallen bottle all require accurate 6D pose tracking to be performed effectively. 


In addition to the full six-dimensional end effector pose tracking, the loco-manipulation problem inherently accounts for three modalities, i.e., locomotion, manipulation, and transition between them, during the long-horizon multi-stage operation.
The dynamic interplay between these modalities presents a significant challenge of achieving seamless coordination among them. Existing approaches such as \cite{Fu2023a,Wang2024,Portela2024,Yokoyama2023} address this challenge by dividing the loco-manipulation into decoupled locomotion and manipulation problems, which do not fully leverage the whole-body coordination capability of the mobile-manipulator robots.


\begin{figure}
    \centering
    \includegraphics[width=\linewidth]{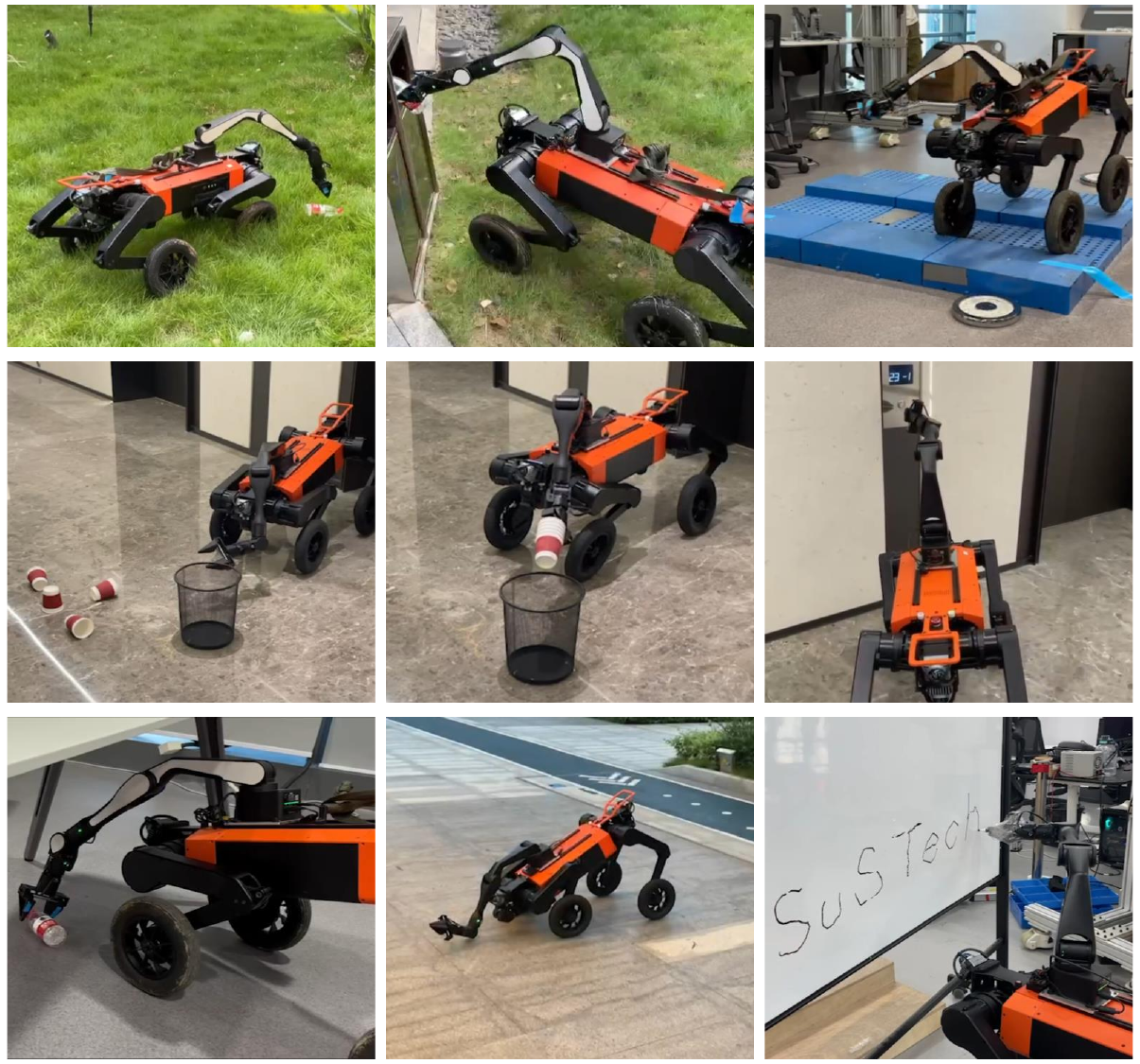}
    \caption{Whole-body 6D task space loco-manipulation for WQM. Writing, outdoor manipulation, picking up cups with random entrance. enter elevators.}
    \label{fig:showcase}
    \vspace{-15px}
\end{figure}

\begin{figure*}[t]
    \centering
    \includegraphics[width=0.9\linewidth]{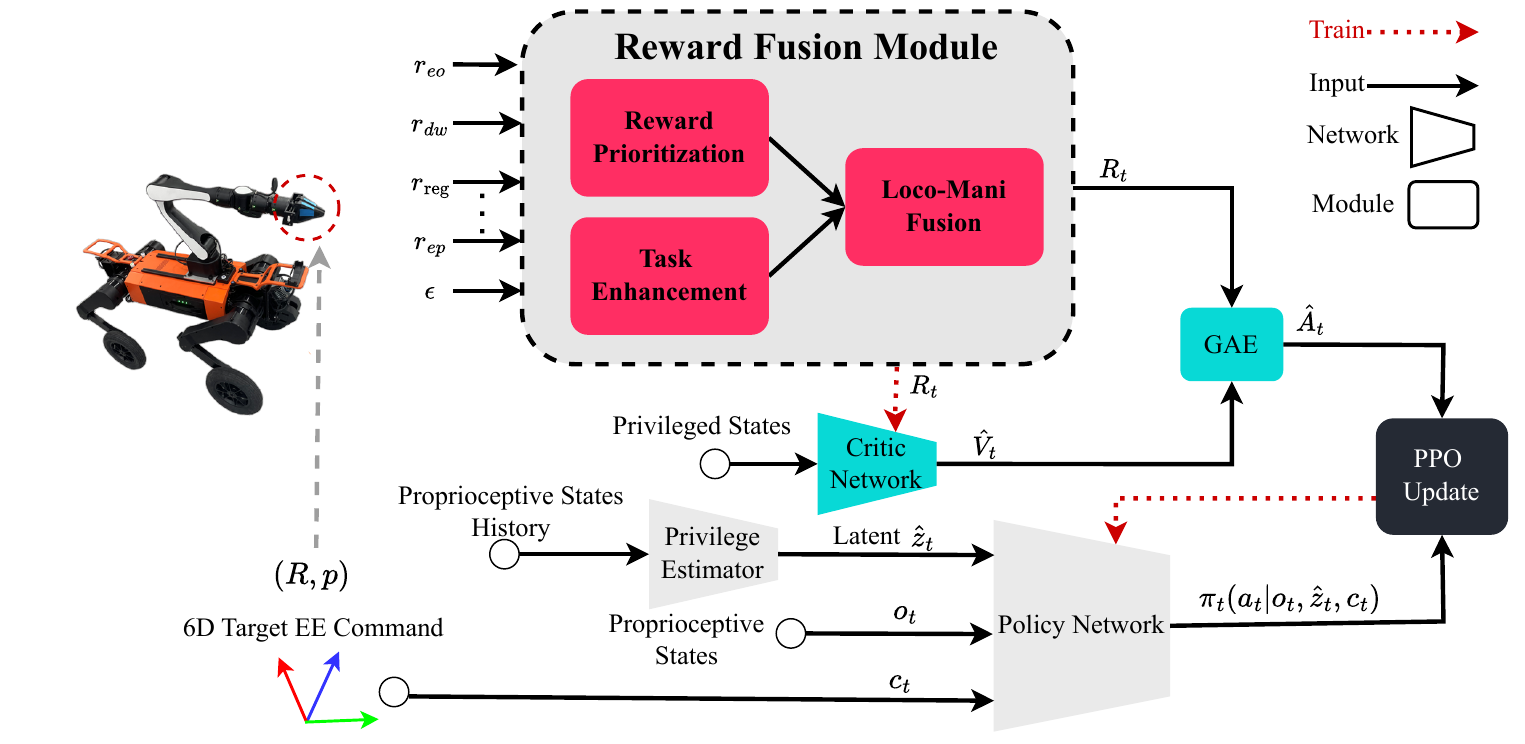}
    \caption{The training paradigm for the whole-body policy using the Reward Fusion Module. Various reward terms are integrated by RFM to generate a reasonable total reward $R_t \in \mathbb{R}$. Then, $R_t$ is sent to generalized advantage estimator (GAE) for Actor training. Moreover, $R_t$ is used for computing TD error to train the Critic. The unified policy takes the 6D target world frame command $(R,p)$ from users, latent $\hat{z}$ encoded from privilege estimator, and proprioceptive states as input, output the whole-body action probability for PPO training.}
    \label{fig: training paradigm}
    \vspace{-0.4cm} 
\end{figure*}

To address the aforementioned challenges, we develop a novel reward fusion module (RFM) to systematically integrate reward terms corresponding to different tasks in a nonlinear manner. In such a way, the inherent multi-stage and hierarchical feature of different modalities in the loco-manipulation problem can be carefully accommodated. By combining the proposed RFM with a teacher-student RL training paradigm, we present a complete RL scheme to achieve 6D EE pose tracking for the wheeled-quadruped manipulator robot. 


Our contributions can be summarized as follows:
\begin{enumerate}
    \item 
    We develop a novel reward fusion module that integrates reinforcement learning reward terms associated with different tasks in a nonlinear manner to achieve nontrivial balance among them. Such a nonlinear reward fusion module allows for simple establishment of task hierarchy and focus prioritization during long horizon tasks.
    \item We have demonstrated the efficacy of the reward fusion module in achieving direct six-dimensional end-effector pose tracking. The RFM module has been shown effective in addressing the direct EE pose tracking problem that naturally requires addressing balance between locomotion task and manipulation task due to the redundant degrees of freedom of the hardware platform. Through incorporating the proposed reward fusion module into a classical teacher-student paradigm, our overall reinforcement-learning based control architecture achieves state-of-the-art loco-manipulation performance in real-world experimental tests.

\end{enumerate}

\section{Related Works}
Recently, the locomotion capabilities of quadrupedal robots have made significant strides through whole-body RL, enabling them to traverse challenging terrains \cite{Jenelten2024a,Hoeller2024,Lee2020a,Yokoyama2023,Ferrolho2023,Arcari2023,Ma2023a}. Incorporation of wheels further enhances quadrupeds' locomotion efficiency, enabling them to move faster and smoothly across flat surfaces. Wheeled quadrupedal robots, which combine the advantages of both legs and wheels, have attracted significant interest due to their unique capabilities and potential \cite{Lee2024,Bjelonic2020,Jelavic2021,Bjelonic2021}. 

The investigation of utilizing RL for whole-body control in loco-manipulation follows the mainstream of the advancement of locomotion, while is still at an early stage. Existing works can be generally categorized into two classes, either decoupled this problem into discrete two modes, or focus on three-dimensional EE position tracking.
\cite{Cheng2023} employs a behavior tree to coordinate between two individual policies taking care of locomotion and manipulation tasks separately, and performs the manipulation task using a leg of the robot. 
Pedipulation \cite{Arm2024} employs a single policy to perform both locomotion and manipulation tasks, using three legs for locomotion and the remaining leg for manipulation. This approach treats loco-manipulation as a unified task without distinguishing between the distinct characteristics of locomotion and manipulation.

\cite{Ma2022a} achieves loco-manipulation with a legged robot by integrating an RL controller for the locomotion task and an MPC controller for the manipulation task, without using the RL controller to account for the whole-body coordination for the overall loco-manipulation problem. 

More recently, \cite{Fu2023a} focuses on achieving the 3D whole-body loco-manipulation. Such a method utilizes a unified policy with two distinct modes to control a quadrupedal manipulator that exhibits whole-body collaboration. These two distinct modes require manual switching, resulting in noticeable pauses during task execution. \cite{Portela2024} focuses on learning a force control policy to explicitly control the EE applied force for legged manipulator. They also learn a loco-manipulation policy, which focuses on tracking base velocity commands and 3D EE target position in body frame. \cite{Wang2024} uses P3O \cite{Zhang2022} to learn a whole-body loco-manipulation policy. Such a method struggles to track the orientation of EE and also requires a base instruction from the user.
In \cite{Liu2024a}, the authors achieved autonomous whole-body loco-manipulation by integrating an inverse kinematics (IK) arm controller with an RL legged controller, all governed by a high-level vision-based planner. As mentioned by the authors, the necessity of using IK stems from the incapability of using RL controller to track the EE orientation. The IK controller in \cite{Liu2024a}'s framework applies the pseudoinverse Jacobian method to solve an unconstrained optimization problem, which neither accounts for joint position/velocity constraints nor incorporates dynamic considerations. Consequently, such a decoupled scheme fails to fully exploit the whole-body capability to track the 6D EE command

\section{Whole-Body Control Pipeline for Loco-Manipulation}
In this paper, we leverage the RL paradigm with RFM to address the whole-body loco-manipulation problem for WQM platform. The problem is to directly track the 6D EE target command that is fixed in world frame using one single policy. As illustrated in Fig~\ref{fig: training paradigm}, our proposed training paradigm contains a unified RL policy, which takes proprioceptive state, privilege latent estimation and 6D task space EE command as inputs. The long history observation is encoded by a privilege estimator to estimate privilege information. 

\subsection{Policy Input}
Diving into the details, as shown in Fig \ref{fig: training paradigm}, our unified policy only needs a 6D target pose command $^BT_{ee}^*\in SE(3)$ from the user.
\begin{equation} \label{se3}
\setlength{\abovedisplayskip}{3pt}
\setlength{\belowdisplayskip}{3pt}
    ^BT_{ee}^* = \begin{bmatrix}
  ^BR_{ee}^* & ^BP_{ee}^*\\
   0&1
\end{bmatrix}, \enspace ^BR_{ee}^*\in SO(3).
\end{equation}
where $^BP_{ee}^*$ is the target EE position w.r.t body frame and $^BR_{ee}^*$ is the target EE orientation w.r.t body frame. Noticeably, the command is fixed w.r.t world frame unless the user changes it. Then the command is represented in body frame since the agent should be agnostic to the inertial frame \cite{Arm2024}. After vectorization, the command is denoted as $\mathbf{c}_t = \text{Vec}(^BR_{ee}^*, ^BP_{ee}^*) \in \mathbb{R}^{12}$. 

The policy input comprises of proprioceptive observation $\mathbf{o}_t \in \mathbb{R}^{81}$, command $\mathbf{c}_t \in \mathbb{R}^{12}$ and privilege latent estimation $\hat{z}_t \in \mathbb{R}^{32}$, as defined in Table~\ref{Table:prorioceptive observation}. The proprioceptive observation $\mathbf{o}_t$ includes joints position of 18 non-wheel joints $q^{nw} \in \mathbb{R}^{18}$, joints velocities $\dot{q} \in \mathbb{R}^{22}$, last action $\mathbf{a}_{t-1} \in \mathbb{R}^{22}$, base angular velocity $^B\omega_b \in \mathbb{R}^{3}$,  projected gravity $^Bg \in \mathbb{R}^{3}$, $SE(3)$ distance reference $\epsilon_t^{\text{ref}} \in \mathbb{R}$ that provides reference for the distance between current EE pose and target EE pose. $\epsilon_t^{\text{ref}}$ will be explained in detail in~\ref{sec:Loco-Manipulation Task and Autonomous Transition}. 
\subsection{Teacher Student Structure}
We construct a teacher-student framework in line with RMA \cite{Kumar2021} to facilitate sim-to-real transfer.
\subsubsection{Teacher training} The teacher is equipped with an MLP privilege encoder, whose input is privilege state including contact forces, friction coefficient, EE twist, etc. Then, the privilege encoder outputs the privilege latent $z \in \mathbb{R}^{32}$, which is taken by teacher policy to generate whole-body action. The privilege encoder is viewed as a part of the teacher policy, and optimized together with teacher policy using PPO.
\subsubsection{Student training} Initially, the student policy copies the parameter of the teacher policy as a warm start. Subsequently, the student MLP privilege estimator learns to approximate $z$ by $\hat{z}$ using the 10 frames of proprioceptive observation history as input. The loss is derived by $ loss =  Mean(\Vert \hat{z} -  sg[z]\Vert^2)$, where $sg[\cdot]$ is the sign of stop gradient. The student policy is fine-tuned with PPO. By using this latent variable representation, the policy can effectively infer and adapt to the underlying dynamic parameters and states, which benefits sim-to-real transfer \cite{Lee2020a}.
\begin{table}[htbp]
\vspace{-1pt}
 \centering
 \caption{Input of Policy}
  \resizebox{0.75\linewidth}{!}{
    \begin{tabular}{ccc}
        \toprule
         Input terms  &  Notations \\
         \midrule
         \rowcolor{lightgray}\multicolumn{2}{|c|}{Proprioceptive Obs. $\mathbf{o}_t$}\\
         Non-wheel joints position &  $q^{nw}\in \mathbb{R}^{18}$ \\ 
         Joints velocity&  $\dot{q} \in \mathbb{R}^{22}$ \\ 
         EE Position&  $^BP_{ee}\in \mathbb{R}^{3}$ \\ 
         EE Orientation & \text{Vec}($^BR_{ee}) \in \mathbb{R}^{9}$   \\ 
         Last action& $\mathbf{a}_{t-1} \in \mathbb{R}^{22}$  \\ 
         Base angular velocity& $^B\omega_b \in \mathbb{R}^{3}$    \\
         Projected gravity    &  $^Bg \in \mathbb{R}^{3}$  \\
        SE(3) distance ref. & $\epsilon_t^{\text{ref}} \in \mathbb{R}$ \\
        \rowcolor{lightgray}\multicolumn{2}{|c|}{Command}\\
        6D EE target & $\mathbf{c}_t\in \mathbb{R}^{12}$ \\
        \rowcolor{lightgray}\multicolumn{2}{|c|}{Privilege latent}\\
        Latent estimation&  $\hat{z}_t \in \mathbb{R}^{32}$ \\
        \bottomrule   
    \end{tabular}}\label{Table:prorioceptive observation}
    \vspace{-1pt}
\end{table}
\subsection{Action Space}
For WQM, since the wheel joints can only be controlled by velocity commands, the output of our policy, $\mathbf{a}_t \in \mathbb{R}^{22}$, is separated into two different categories, consisting of the target positions for non-wheel joints $a_t^{nw}$ and the target velocities for wheel joints $a_t^{w}$. These outputs are then clipped by user-appointed torque limits. After the clip, $\mathbf{a}_t$ is sent to a separate Proportional-Derivative (PD) controller, which generates whole-body torques $\tau_t = (\tau_t^{nw},\tau_t^{w}) \in \mathbb{R}^{22}$ for the WQM,
\begin{equation}
\setlength{\abovedisplayskip}{3pt}
\setlength{\belowdisplayskip}{3pt}
    \begin{aligned}
        \tau^{nw}_t &= K_p^{nw}(\mathbf{a}^{nw}_t - q_{n}+ q_t) - K_d^{nw}\dot{q}_t^{nw},\\
        \tau^{w}_t &= K_d^w(\mathbf{a}^w_t - \dot{q}_t^w).
    \end{aligned}
\end{equation}
where the superscript $nw$ denotes non-wheel joints, $w$ denotes wheel joints. $K_p,K_d$ are the PD parameters. $q_{n}$ is the nominal joint position and $q_{t}$ is the measured joint position at time $t$.

Our unified policy is trained with standard PPO \cite{Schulman2017} combined with a novel RFM structure as shown in Fig~\ref{fig: training paradigm}. RFM provides a systematic way to perform reward fusion in a non-linear manner. Typically, all the rewards items are merged into one value using weighted sum, which is fundamentally a linear combination. In view of Fig~\ref{fig: training paradigm},there are three key aspects in RFM, Task Enhancement, Reward Prioritization, and Loco-mani fusion. In the following section, we illustrate the proposed RFM.

\section{RFM: Reward Fusion Module} \label{section: Reward Design}
In this paper, we leverage a novel structure named Reward Fusion Module (RFM) to address the 6D EE pose tracking problem for WQM. In reinforcement learning, reward design plays a central role in the success of valid controller synthesis. Generally speaking, the reward design process first aims to design a set of individual reward terms $\{r_1,r_2,...r_n\}$ associated with sub-tasks for the overall objective. Then, all individual reward terms are integrated together as the final reward. One of the most widely adopted way of integrating individual reward terms is through weighted sum $r = \omega_1r_1 + \omega_2r_2 + ... + \omega_nr_n$, which is essentially a linear combination of all sub-task rewards.  Such a linear reward integration technique makes it difficult to reflect complex objectives such as hierarchical tasks or multi-stage task. To address this challenge, we develop a novel reward fusion module that offers a nonlinear reward integration scheme to achieve complex balance among all reward terms.

\subsection{RP: Reward Prioritization}

For loco-manipulation problem, when the target EE position is distant, tracking of target EE orientation is usually not important and potentially increase the failure rate and energy consumption. Only when the target pose is within the workspace of the arm does orientation tracking make sense. In essence, there is an inner priority between position tracking and orientation tracking during the long-horizon loco-manipulation task, which calls for nonlinear reward fusion techniques. Inspired by the relationship between Hierarchical Whole-Body Control (HWBC)~\cite{Bellicoso2016} and Weighted Whole-Body Control (WWBC)~\cite{Grandia2023a} that deal with whole-body coordination from the model-based perspective, we propose reward prioritization scheme to achieve such task hierarchy in our RL framework.

In RFM, we design two reward terms concerning EE position tracking $\mathbf{r}_{ep} \in (0,1]$ and EE orientation tracking $\mathbf{r}_{eo} \in (0,1]$.
The priority between $\mathbf{r}_{ep}$ and $\mathbf{r}_{eo}$ is established by
\begin{equation}
    \label{eq: priority_r_ep_r_eo}
    \begin{aligned}
    \mathbf{r}_t = \mathbf{r}_{ep} + \mathbf{r}_{ep}\mathbf{r}_{eo}, \quad
    \frac{\partial\mathbf{r}_t }{\partial\mathbf{r}_{ep}} = 1+\mathbf{r_{eo}}, \quad
    \frac{\partial\mathbf{r}_t }{\partial\mathbf{r}_{eo}} = \mathbf{r}_{ep}
    \end{aligned}
\end{equation}
Considering the partial derivative of $\mathbf{r}_t$ as shown in \eqref{eq: priority_r_ep_r_eo}, $\mathbf{r}_{eo}$ only takes effect when $\mathbf{r}_{ep}$ is sufficient, meanwhile $\mathbf{r}_{ep}$ always have impact on $\mathbf{r}_t$. $\mathbf{r}_t$ reaches its maximum if both $\mathbf{r}_{ep}$ and $\mathbf{r}_{ep}$ equal one. This is exactly the priority we want to achieve. Moreover, there is no any parameter to be fine-tuned for establishing such a priority.

In RFM, user can designate priority between different tasks and rewards. This module is called Reward Prioritization (RP). RP guides the achievement of hierarchical objectives by establishing priorities among rewards, simply through multiplication. The priority between the EE position and orientation tracking is easily established without fine-tuning process. Here, the specific forms of $\mathbf{r}_{ep}$ and $\mathbf{r}_{eo}$ are defined as follows,
\begin{equation} \label{eq: r_ep and r_eo before enhacne}
    \begin{aligned}
        \mathbf{r}_{ep} = e^{-d_p/\sigma}, \enspace
        \mathbf{r}_{eo} = e^{-d_{\theta}/\sigma_s}
    \end{aligned}
\end{equation}
where $d_{\theta} = ||\log(R_{ee}^*R_{ee}^T)||_F$ is the Frobenius norm of the logarithm of the special orthogonal group SO(3) and $ d_p = ||P_{ee}^*-P_{ee}||_2$ is Euclidean distance~\cite{Park1995}.

The second priority is to ensure that the robot maintains desired motion while approaching the target without falling down. To achieve this, we first shape a reward item $\mathbf{r}^{\text{mani}}_{\text{reg}}$ that regularize the robot to retain smooth and feasible movements throughout the manipulation process. Then, RP is utilized to prioritize $\mathbf{r}^{\text{mani}}_{\text{reg}}$ over tracking-related rewards as follows
\begin{equation}
    \mathbf{r}_t = \mathbf{r}^{\text{mani}}_{\text{reg}} + \mathbf{r}^{\text{mani}}_{\text{reg}} (\mathbf{r}_{ep} + \mathbf{r}_{ep}\mathbf{r}_{eo})
\end{equation}
By prioritizing regularization, the agent will not choose to take risks of failure to get a higher tracking-related reward, since tracking-related reward only takes effect when $\mathbf{r}^{\text{mani}}_{\text{reg}}$ is guaranteed.

The specific meaning of ``regularization" varies according to the requirements of users and tasks. For examples, in flat terrain locomotion task, ``regularization'' may mean stable base height and flat base orientation, which is not true for manipulation task. Because in manipulation task, the base need to bend or stretch to cooperate with the manipulator. Therefore, our customized definition of ``regularization" is omitted.

RP can generalize to other hierarchical tasks. For example, most of other approaches include two kinds of commands, which are base velocity command $\mathbf{r}_{bv}$ and EE pose command that is fixed in body frame. For achieving such a demand, just design the fusion process as follows
\begin{equation}
    \mathbf{r}_t =  \mathbf{r}^{\text{mani}}_{\text{reg}} +\mathbf{r}^{\text{mani}}_{\text{reg}}( \mathbf{r}_{ep} + \mathbf{r}_{ep}\mathbf{r}_{eo}) + \mathbf{r}^{\text{loco}}_{\text{reg}}\mathbf{r}_{bv}
\end{equation}


\subsection{Enhancement: Precise 6-D Pose Tracking} \label{Achieving Precise Tracking}
Tracking accuracy is one of the most important metrics in evaluating loco-manipulation tasks, which is even more significant in 6D EE pose tracking for precise manipulation. Our enhancement module is tailored to improve tracking accuracy, which consists of two parts, micro Enhancement and cumulative-based error penalty, respectively.

\subsubsection{Micro Enhancement}

\begin{figure}[htbp]
\vspace{-1pt} %
    \centering
    \includegraphics[width=0.95\linewidth]{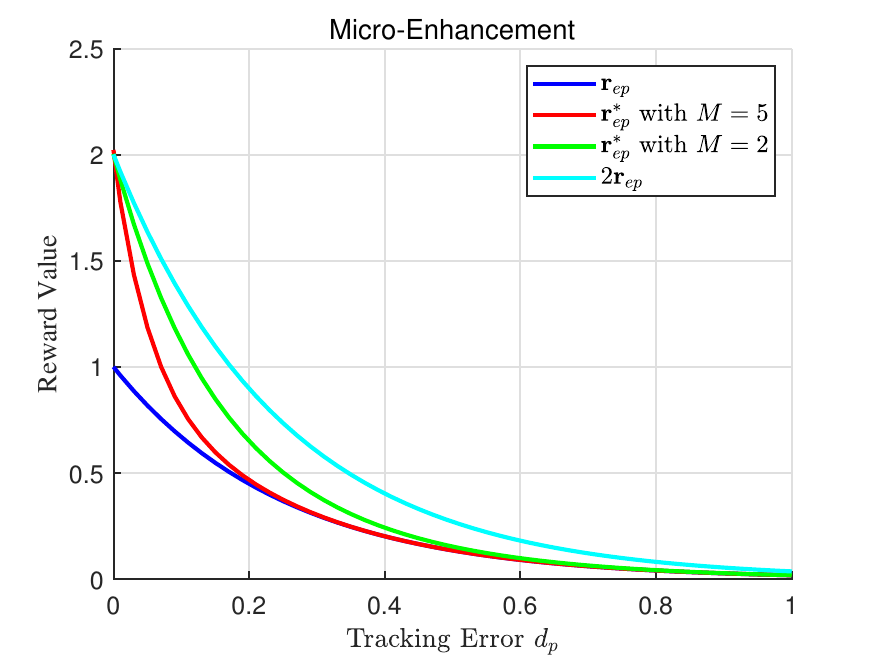}
  \caption{Illustration of Micro-Enhancement}
  \label{fig:enhance}
  \vspace{-1pt} %
\end{figure}

To encourage accurate EE position tracking , one heuristic is to increase the weight assigned to $\mathbf{r}_{ep}$ in \eqref{eq: r_ep and r_eo before enhacne}. However, increasing the weight will affect the reward value at all tracking error range, and potentially making other reward terms less influential. As a result, we only want to increase the marginal reward value of $\mathbf{r}_{ep}$, which is achieved by the here-proposed micro enhancement scheme.
In RFM, $\mathbf{r}_{ep}$ and $\mathbf{r}_{eo}$ are micro-enhanced as follows:
\begin{equation} \label{r_ep and r_eo}
\setlength{\abovedisplayskip}{3pt}
\setlength{\belowdisplayskip}{3pt}
    \begin{aligned}
        &\mathbf{r}_{ep}^* = \mathbf{r}_{ep} + (\mathbf{r}_{ep})^M \quad
        \mathbf{r}_{eo}^* = \mathbf{r}_{eo} + (\mathbf{r}_{eo})^M
    \end{aligned}
\end{equation}
where $M>1$ is the micro enhancement parameter influencing the shape of curve of $\mathbf{r}_{ep}^*$. The proposed enhancement term $(\mathbf{r}_{ep})^M$ takes the $M$-th power of  $\mathbf{r}_{ep}$, which essentially enlarges the reward gradient when tracking error is relatively small, while not influencing the gradient when tracking error is large. We plot a figure to intuitively illustrate the micro enhancement using $\mathbf{r}_{ep} = e^{-d_p/ 0.25}$ as shown in Fig~\ref{fig:enhance}.

\subsubsection{Cumulative Penalty Mechanism} \label{cumulative error penalty} In traditional PID control, the integral portion is usually used to eliminate steady-state error and improve control accuracy. Inspired by potential-based reward design \cite{Jeon2023a} and the principles of PID control, we develop a cumulative penalty mechanism to emulate the functionality of integral portion of PID controller. 
The cumulative error and cumulative-based penalty are defined as follows,
\begin{equation}\label{r_cb}
\setlength{\abovedisplayskip}{3pt}
\begin{aligned}
    \mathbf{r}_{cb,t} = e^{cb}_t, \qquad
    e^{cb}_{t} = e^{cb}_{t-1} + \kappa\cdot\epsilon_t
\end{aligned}
\end{equation}
where $\kappa$ is the weight scalar. $\epsilon_t$ is the tracking error at time $t$ introduced by a standard criterion $f:SE(3)\times SE(3)\longrightarrow\mathbb{R}^+$ \cite{Park1995}
\begin{equation} \label{eq:SE3 distance}
\begin{aligned}
    \epsilon_t =& f(T^*_{ee}, T_{ee,t})\\
    =&a_1||\log(R_{ee,t}^*R_{ee}^T)||_F + a_2||P_{ee,t}^*-P_{ee}||_2 
\end{aligned}
\end{equation}
where $a_1,a_2$ are weighting parameters. In practice, A clipping function is implemented to limit the upper bound of $e^{cb}$. Otherwise, excessively large $e^{cb}$ may result in system overreaction.

We hypothesize that the cumulative error penalty helps the agent overcome the local minima. If the agent becomes stuck in such a local minimum, it will receive an ever-increasing penalty, causing the value function at that state to continually decrease. This mechanism drives the agent to precisely reach the target command. Moreover, it is noticeable that $\mathbf{r}_{cb}$ essentially contains historical information, making it essential to include $e^{cb}$ at least as a privilege information inputted into the Critic.

\subsection{Loco-Mani Fusion: Achieving Smooth Loco-Manipulation}\label{sec:Loco-Manipulation Task and Autonomous Transition}

In fact, loco-manipulation task is one of the multi-stages tasks. There are two stages that need to be balanced, which are locomotion and manipulation. Most of previous works achieve this balance in a discrete manner, which results in non-smooth task executions.

In this paper, we develop a whole-body policy that unifies locomotion and manipulation into smooth loco-manipulation. To achieve this, RFM first divides the loco-manipulation task into three separate reward groups as shown in Tab~\ref{Table:Reward Design}. \textbf{Locomotion rewards}: $\mathbf{r}_{\text{loco}}$, 
\textbf{Manipulation rewards}: $\mathbf{r}_{\text{mani}}$ and \textbf{Basic rewards}: $\mathbf{r}_{\text{basic}}$. 
Then, we introduce a phase variable $\mathcal{D}(\epsilon_t^{\text{ref}})$ to handle the reward fusion of these 3 categories.
\begin{equation} \label{eq:r_t}
\setlength{\abovedisplayskip}{3pt}
\setlength{\belowdisplayskip}{3pt}
    \begin{aligned}
        &\mathbf{r}_t = (1-\mathcal{D}(\epsilon_t^{\text{ref}})) \cdot \mathbf{r}_{\text{mani}}+\mathcal{D}(\epsilon_t^{\text{ref}}) \cdot \mathbf{r}_{\text{loco}} + \mathbf{r}_{\text{basic}}
    \end{aligned}
\end{equation}
where $\mathcal{D}(\epsilon_t^{\text{ref}})$ is dependent upon the $SE(3)$ distance reference \(\epsilon_t^{\text{ref}}\). $\epsilon_t^{\text{ref}}$ is defined as
\begin{equation} \label{eq:SE3 distance ref}
    \epsilon_{t}^{\text{ref}} = \max\left\{\epsilon_0 - v t, 0 \right\}
\end{equation}
where the $\epsilon_{t}^{\text{ref}}$ is the SE(3) distance reference at time $t$. $\epsilon_0$ is SE(3) distance at initial state following the definition of \eqref{eq:SE3 distance}. $v_{min}<v<v_{max}$ is the user-appointed decrease velocity of the SE(3) distance reference. In fact, $v$ determines how fast the agent move to track the 6D EE target.
\begin{table}[htbp]
\vspace{-1pt}
 \centering
 \caption{Summary for Reward Shaping}
 \resizebox{\linewidth}{!}{\begin{tabular}{ccc}
        \toprule
         Reward Term &  Equation & Weight\\ 
         \midrule
         \rowcolor{lightgray} \multicolumn{3}{|c|}{Manipulation Reward $\mathbf{r}_{\text{mani}}$}\\
        Tracking EE position & $\mathbf{r}_{ep} = e^{-d_p/\sigma}$ & / \\ 
         Tracking Ee orientation  &  $\mathbf{r}_{eo} = e^{-d_{\theta}/\sigma}$ & / \\
         Cumulative-based & $\mathbf{r}_{cb} = \min\left\{e^{cb}, 20\right\}$  & / \\ 
         Potential-based \cite{Jeon2023a} & $\mathbf{r}_{pb}$ &  / \\ 
        All contact& $\mathbf{r}_{ac} = 1 \text{ if all feet grounded else } 0$ & / \\
        Reg. of mani.&  $\mathbf{r}_{\text{reg}}^{\text{mani}}$ & /\\
        \rowcolor{lightgray} \multicolumn{3}{|c|}{Locomotion Reward $\mathbf{r}_{\text{loco}}$}\\
        Displacing WQM& $\mathbf{r}_{dw,t} = \exp\left(- e_t^{\text{ref}} / \sigma_s \right)$ & /\\ 
        Static arm & $\sum_{i \in \mathcal{A}}\vert q_i - q_{i,sa}\vert $   & -0.15 \\
        Reg. of loco.&  $\mathbf{r}_{\text{reg}}^{\text{loco}}$ & / \\
        \rowcolor{lightgray} \multicolumn{3}{|c|}{Basic Reward $\mathbf{r}_{\text{basic}}$}\\
        Action rate & $\left|a_{t-1}-a_{t}\right|^{2}$& -0.003\\
        Action smoothness & $\left|a_{t-2}-2 a_{t-1}+a_{t}\right|^{2}$ & -0.001\\
        Collision & $1\enspace\text{if collides else} \enspace0$ & -5\\
        Alive & $1\enspace\text{if alive else} \enspace0$ & 2.0\\
        Weighted Power & $\sum \Vert\tau \cdot \dot{q}\Vert \cdot \Omega_{p}$ & $-3.3\times10^{-4}$\\
        Weighted Torque & $\sum \tau^2  \cdot \Omega_{\tau}$ & $-4\times 10^{-5}$\\
        Relative Torques rate& $\sum \frac{\Vert\tau_t - \tau_{t-1}\vert}{ \tau_{\text{limit}}}$ & $-0.1$\\
         
        \bottomrule 
        
    \end{tabular}}
    \label{Table:Reward Design}
    \vspace{-10pt}
\end{table} 

$\mathcal{D}(x)$ is a sigmoid function, which is 
\begin{equation} \label{eq:sigmoid_scale}
    \begin{aligned}
       \mathcal{D}(x;\mu,l) &= \frac{1}{1+e^{-5\cdot(x-\mu)/{l}}} \\
    \end{aligned}
\end{equation}
where $\mu$ is the middle value of the curve and $l$ serves as the slope of the curve. 

\begin{figure}[htbp]
\vspace{-1pt} 
    \centering
    \includegraphics[width=0.9\linewidth]{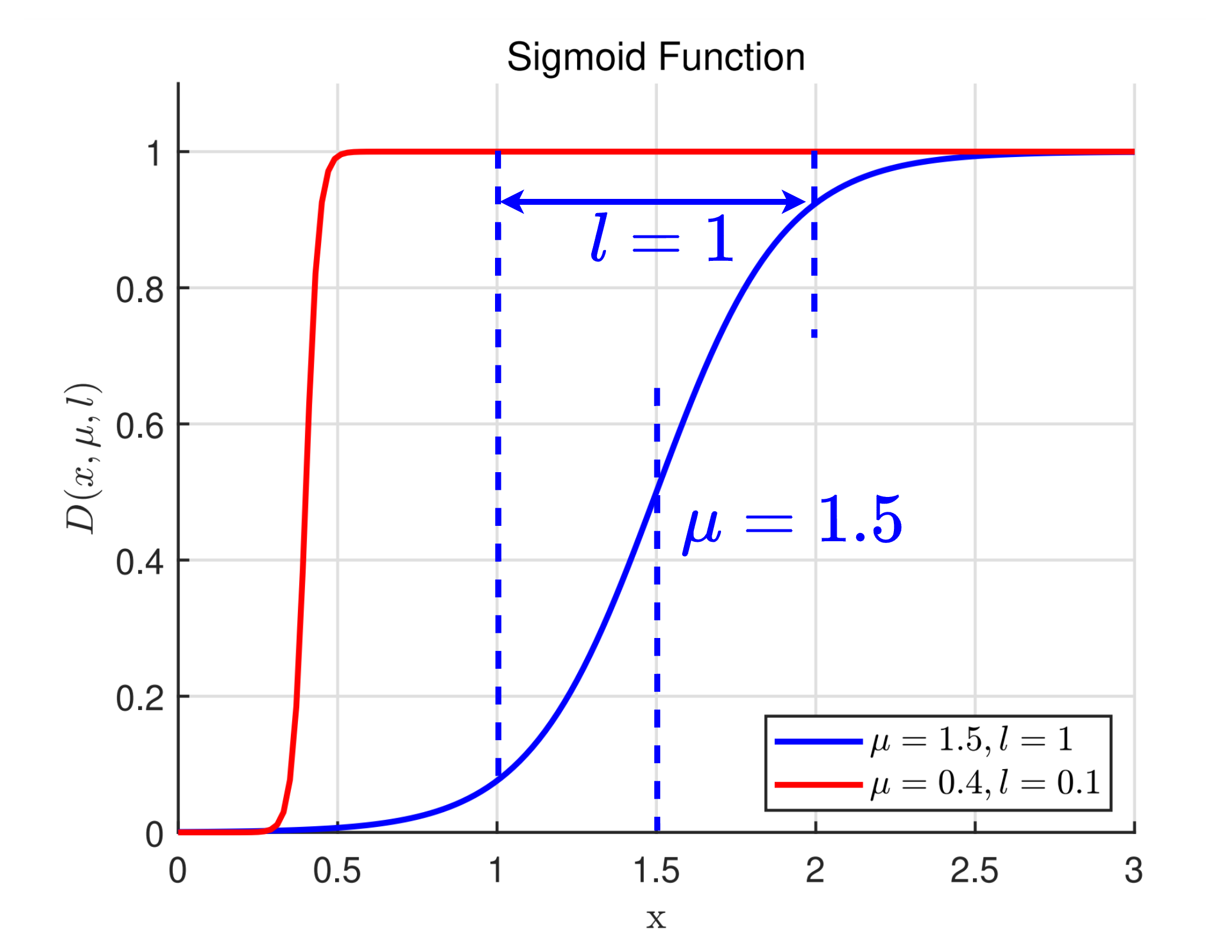}
  \caption{The curve of $\mathcal{D}(x,\mu,l)$ with two different parameters setting}
  \label{fig:sigmoid_scale}
  \vspace{-5pt} 
\end{figure}

For a more intuitive understanding, the curve of \eqref{eq:sigmoid_scale} is shown in Fig~\ref{fig:sigmoid_scale} with two different parameters setting. The parameter-tuning of $\mu$ and $l$ is very simple, since it follows the principle of armspan. Mathematically, given that the arm span is $l_a$, we recommend to set $\mu = 2l_a,\quad l=2l_a$.

$\epsilon_t^{\text{ref}}$ induces the robot movement to the target EE pose at an even speed. Naturally, the reward of displacing WQM , $\mathbf{r}_{dw}$, is introduced: 
\begin{equation} \label{eq: r_dw}
        \begin{aligned}
            e^{\text{ref}}_t &= \max\left\{\vert \epsilon_{t}^{\text{ref}} - \epsilon_t\vert - \gamma, 0\right\} \\
            \mathbf{r}_{dw,t} &= \exp\left(- e_t^{\text{ref}} / \sigma_s \right) 
        \end{aligned}
    \end{equation}
where $\sigma_s$ is the parameter. $\mathbf{r}_{dw}$ is included as one part of $\mathbf{r}_{\text{loco}}$, acts as a coarse reference for the locomotion process. The release parameter $\gamma$ eases the tracking accuracy requirement in the middle of tracking process.

It is notable that $\epsilon_t^{\text{ref}}$ mentioned in \eqref{eq:SE3 distance ref} only depends on the initial state and time. With $\mathcal{D}(\epsilon_{t}^{\text{ref}})$, the WQM learns to perform smooth and stable motion according to the user-appointed $v$. 
The introduction of $\mathcal{D}(\epsilon_t^{\text{ref}})$ successfully integrates locomotion with manipulation, achieving smooth loco-manipulation with dynamic focus based on the value of $\epsilon_t^{\text{ref}}$. Noticeably, the parameter $\kappa$ in \eqref{r_cb} is selected as $\kappa = (1-\mathcal{D}(\epsilon_t^{\text{ref}}))$, where $\mathcal{D}(\epsilon_t^{\text{ref}})$ is defined in \eqref{eq:SE3 distance ref}


\subsection{Summery of RFM}
Summarizing the methodology we discussed above, we can outline the reward design as follows
\begin{equation} \label{eq: r_mani}
\setlength{\abovedisplayskip}{3pt}
\setlength{\belowdisplayskip}{3pt}
\begin{aligned}
   &\mathbf{r}_{\text{mani}} =  {r}^{\text{mani}}_{\text{reg}} +  \mathbf{r}^{\text{mani}}_{\text{reg}}\left(\mathbf{r}^*_{ep} +\mathbf{r}_{ep}\mathbf{r}_{eo}^*\right) + \mathbf{r}_{pb} - \mathbf{r}_{cb} + \mathbf{r}_{ac}\\
   &\mathbf{r}_{\text{loco}} = \mathbf{r}^{\text{loco}}_{\text{reg}} + \mathbf{r}^{\text{loco}}_{\text{reg}} \mathbf{r}_{dw} - \omega_{sa}\mathbf{r}_{sa}  \\
   &\mathbf{r}_t = (1-\mathcal{D}(\epsilon_t^{\text{ref}})) \cdot \mathbf{r}_{\text{mani}}+\mathcal{D}(\epsilon_t^{\text{ref}}) \cdot \mathbf{r}_{\text{loco}} + \mathbf{r}_{\text{basic}}
\end{aligned}
\end{equation}                          

It is worth mentioning that, at the reward fusion level where all individual reward terms are already given, there are no weights to adjust to achieve proper balance between each reward item in $\mathbf{r}_{\text{mani}}$. Furthermore, there is only one weight to be tuned in $\mathbf{r}_{\text{loco}}$.

\section{Experiments and Results with WQM} \label{Experiments}

\subsection{Training and Platform Setup}
To train the RL policy, we use PPO \cite{Schulman2017} combined with SAC \cite{Haarnoja2018} to train our policy $\pi_\theta$. We use Isaac Gym \cite{Makoviychuk2021} for RL training and utilize Mujoco \cite{Todorov2012} for sim-to-sim deployment, since Mujoco's physics simulation is closer to the real world than Isaac Gym. We adopt the idea of Curriculum Learning \cite{Narvekar2020} and Decaying Entropy Coefficient \cite{Lee2023a}. We add randomization to the terrain with a height field akin to \cite{Arm2024}. We add domain randomization during training to bridge the sim-to-real gap. Our agent is trained with 4096 parallel environments based on Legged Gym \cite{Rudin2022}. We designate 7000 iterations with 24 environment steps per iteration for both teacher policy training, and 4000 iterations for the student to be fine-tuned. Our training is conducted on the NVIDIA RTX3090 and approximately demands 8 hours of training duration for teacher training and 4 hours student training.

\begin{figure}[htbp]
    \centering
    \vspace{-1pt} 
    \includegraphics[width=0.9\linewidth]{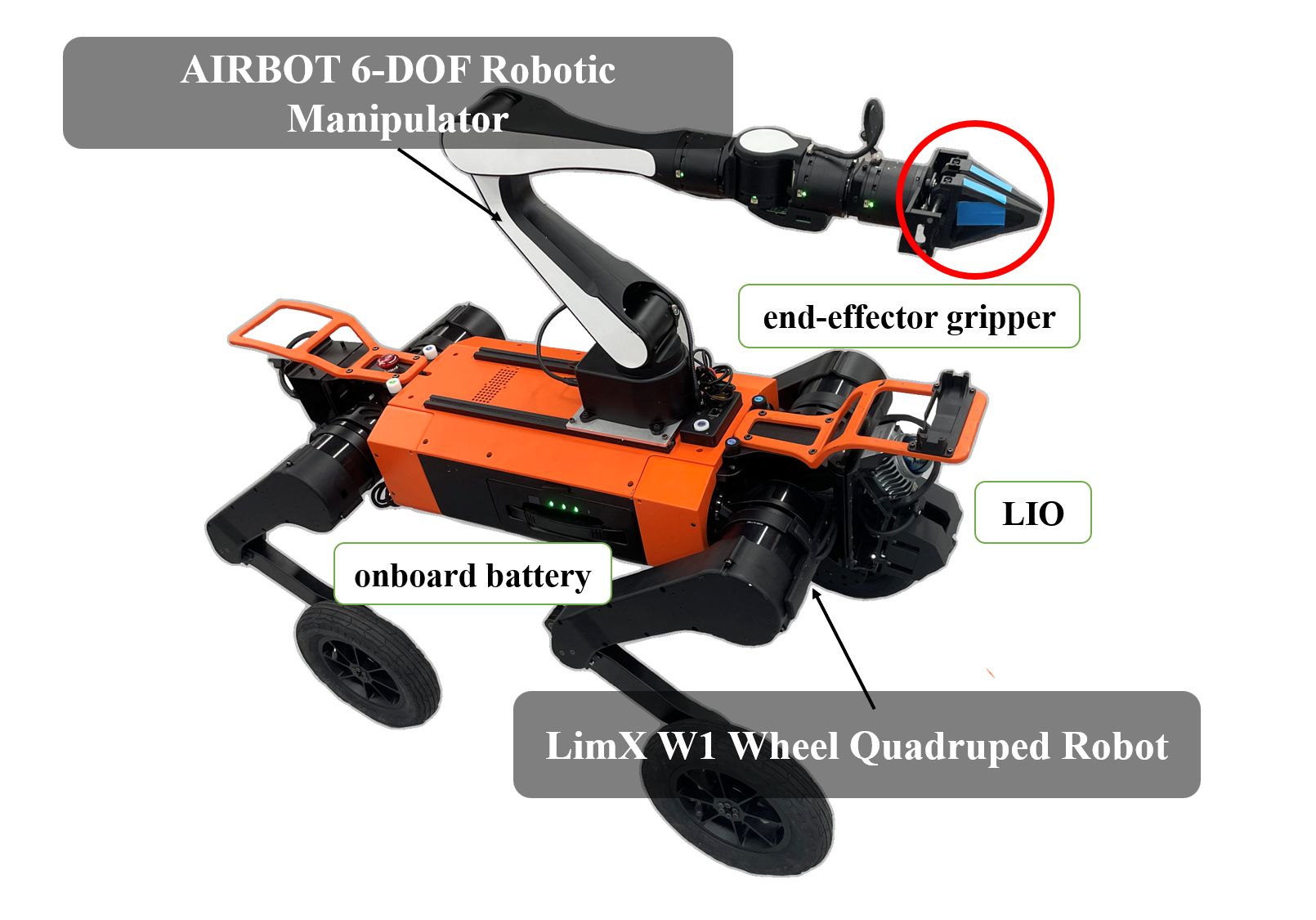}
  \caption{WQM setup}
  \label{fig: WQM setup}
  \vspace{-1pt} 
\end{figure}

The WQM consists of LimX W1, a wheeled quadruped robot equipped with 16 torque-controlled actuators, and a AIRBOT 6-DoF torque-controlled arm, as shown in Fig~\ref{fig: WQM setup}. The power supply for both the W1 and AIRBOT comes from the onboard battery of the W1. The applied torque for both the wheeled quadrupedal robot and the arm is managed by a unified neural network policy that performs inference on the onboard computer of W1. To track the EE 6D world frame target in an outdoor setting, the base position in world frame is estimated by Lidar Inertial Odometry (LIO). Then, the forward kinematics is utilized to compute the EE position in world frame.


\subsection{Simulation: Ablation Studies}

We conducted a series of ablation studies (AS) in a simulated environment to quantitatively assess the effectiveness of the proposed RFM. We report the followwing metrics: 1. success rate defined as \eqref{eq:success rate}. 2. EE position error $d_p$ and EE orientation error $d_{\theta}$ defined in \eqref{eq: r_ep and r_eo before enhacne}. 3. Average Power consumption for all joints. 4. Average joint acceleration for all joints. 5. Nominal deviation defined as \eqref{eq: nominal deviation}

    Success Rate is defined as follows:
    \begin{align} \label{eq:success rate}
        \text{Success Rate} &= \frac{N - F}{N} \times 100\%
    \end{align}
    where \(N\) is the total number of experiments and \(F\) is the number of failures. An agent is considered to have failed in an experiment if any of the following conditions are met:
    \begin{itemize}
        \item Z-component of the projected gravity is larger than -0.1.
        \item The base height is lower than 0.2 m.
        \item The end effector (EE) or the base is in collision.
    \end{itemize}
    
    
    Nominal Deviation is defined by:
    \begin{align} \label{eq: nominal deviation}
        p_{\text{devi.}} = \frac{1}{T} \sum_{j=1}^{T} \sum_{i \in \text{wheel links}} \vert p_{i} - p_{n,i} \vert
    \end{align}
    where \(p_i \in \mathbb{R}^{2}\) is the xy component of i-th wheel' position w.r.t the body frame, and \(p_{n,i} \in \mathbb{R}^{2}\) is the nominal xy position of the i-th wheel that is fixed in the body frame. which serves as reference to measure the quality of the foothold placement. \(T\) is the number of environment steps in one experiment.

We conducted 1000 experimental trials in the simulation for each ablation study policy to collect data. The end effector targets were randomly resampled around the robot, maintaining consistent experimental conditions across all ablation policies. Failure trials were excluded from the dataset when calculating all metrics, except for the Success Rate. The following section outlines the differences among each ablation study policy.

\begin{itemize} 
     \item (RFM w/o loco-mani fusion) Specifically, in this AS, we set
     \begin{equation}
         \mathbf{r}_t = \omega_1 \mathbf{r}_{\text{mani}}+ \omega_2 \mathbf{r}_{\text{loco}} + \omega_3 \mathbf{r}_{\text{basic}}
     \end{equation}
     The optimal parameters we obtain during a two-day tuning are $\{\omega_1 = 1.2 ,\omega_2 = 0.4,\omega_3 = 1.0\}$.
     \item (RFM w/o Reward Prioritization) Specifically, in this AS, $\mathbf{r}_{\text{mani}}$ is changed into 
     \begin{equation} \label{eq: r_mani_without_RP }
        \setlength{\abovedisplayskip}{3pt}
        \setlength{\belowdisplayskip}{3pt}
        \begin{aligned}
       \mathbf{r}_{\text{mani}} = &\omega_{1}\!\cdot \!\mathbf{r}^*_{ep} +\omega_{2}\!\cdot\!\mathbf{r}_{eo}^*
       +\omega_{3}\cdot\mathbf{r}_{pb}\\  &+\omega_{4}\cdot\mathbf{r}_{cb}
        + \omega_5\cdot\mathbf{r}^{\text{mani}}_{\text{reg}} + \omega_6\cdot\mathbf{r}_{ac}
    \end{aligned}
    \end{equation}
     Different from \eqref{eq: r_mani}, we utilize weighted sum instead of RP. We suppose that it can more vividly showcase the superior of RP if we tune massive weights with significant effort. We select $\omega_{1} = 2.0, \omega_{2} = 3.0, \omega_{3} = 3, \omega_{4} = -0.6, \omega_{5} = 3.0, \omega_{6} = 1.5$. They are the best among tuning for 48 hours.
     \item (RFM w/o Enhancement) Specifically, This means that $\mathbf{r}_{\text{mani}}$ is changed into 
     \begin{equation}
         \mathbf{r}_{\text{mani}} =  \mathbf{r}^{\text{mani}}_{\text{reg}} +  \mathbf{r}^{\text{mani}}_{\text{reg}}\left(2\mathbf{r}_{ep} +2\mathbf{r}_{ep}\mathbf{r}_{eo}\right) + \mathbf{r}_{pb} + \mathbf{r}_{ac}
     \end{equation}
     This experiment showcases the improvement of tracking accuracy brought by Enhancement.
     \item (Standard PPO w/o RFM) This is the baseline of our ablation study. The $\mathbf{r}_t$ is designed as
     \begin{equation} \label{eq: r_mani without RFM}
    \setlength{\abovedisplayskip}{3pt}
    \setlength{\belowdisplayskip}{3pt}
    \begin{aligned}
       &\mathbf{r}_{\text{mani}} =  \omega_1\mathbf{r}^{\text{mani}}_{\text{reg}} +  \omega_2\mathbf{r}_{ep} +\omega_3\mathbf{r}_{eo} + \omega_4\mathbf{r}_{pb} + \omega_5\mathbf{r}_{ac}\\
       &\mathbf{r}_t = \omega_6 \mathbf{r}_{\text{mani}}+ \omega_7 \mathbf{r}_{\text{loco}} + \omega_8 \mathbf{r}_{\text{basic}}
    \end{aligned}
     \end{equation}
     The optimal parameters we obtain during a two-day tuning are listed as follows.  $\omega_1=1.0,\omega_2=2.0,\omega_3 = 3.0,\omega_4 = 3.0,\omega_5=1.0,\omega_6 = 1.2,\omega_7=0.5,\omega_8 = 1.0$.
 \end{itemize}

\subsubsection{(RFM w/o Loco-mani Fusion)}
We evaluate the impact of removing the Loco-mani Fusion from our RFM framework and try to balance the locomotion and manipulation task by ``weighted sum". This absence led to a failure issue and jittery motion reflected by excessive average joint acceleration and average power consumption. Without Loco-mani Fusion, the robot cannot properly and smoothly switch focus between locomotion and manipulation task.

\begin{table*}[htbp]
    \centering
    \caption{Ablation study results for our method, without Loco-Mani Fusion, without RP, without enhancement and without RFM.}
    \begin{tabular}{lcccccc}
        \toprule
        & \multicolumn{1}{c}{Success Rate $\uparrow$} 
        & \multicolumn{1}{c}{Aver. Power (W) $\downarrow$} 
        & \multicolumn{1}{c}{Aver. Acc. (rad$\cdot$s$^{-2}$) $\downarrow$} 
        & \multicolumn{1}{c}{EE pos. error (m) $\downarrow$} 
        & \multicolumn{1}{c}{EE ori. error (rad) $\downarrow$} 
        & \multicolumn{1}{c}{Nominal Devi. (m) $\downarrow$} 
        \\
        \midrule
        Ours & $\textbf{99.0}\%$ & $70.983$ & $143.747$ & $\textbf{0.022} $ & $0.041 $ & $\textbf{0.115}$ \\
        w/o LMF & $82.1\%$ & $122.923$ & $332.884$ & $0.023 $ & $0.066 $  & $0.228$ \\
        w/o RP & $93.0\%$ & $65.321$ & $171.341$ & $0.045 $ & $\textbf{0.029} $ & $0.393$ \\
        w/o En. & $98.7\%$ & $\textbf{62.091}$ & $\textbf{121.317}$ & $0.036 $ & $0.117 $ & $0.117$ \\
        w/o RFM & $92.9\%$ & $87.223$ & $ 186.940$ & $0.168 $ & $0.030 $ & $0.467$ \\
        \bottomrule
    \end{tabular}
    \label{tab:ablation study}
\end{table*}
\begin{figure*}[htbp]
\begin{minipage}{0.25\linewidth}
    \centerline{\includegraphics[height=3cm,width=\linewidth]{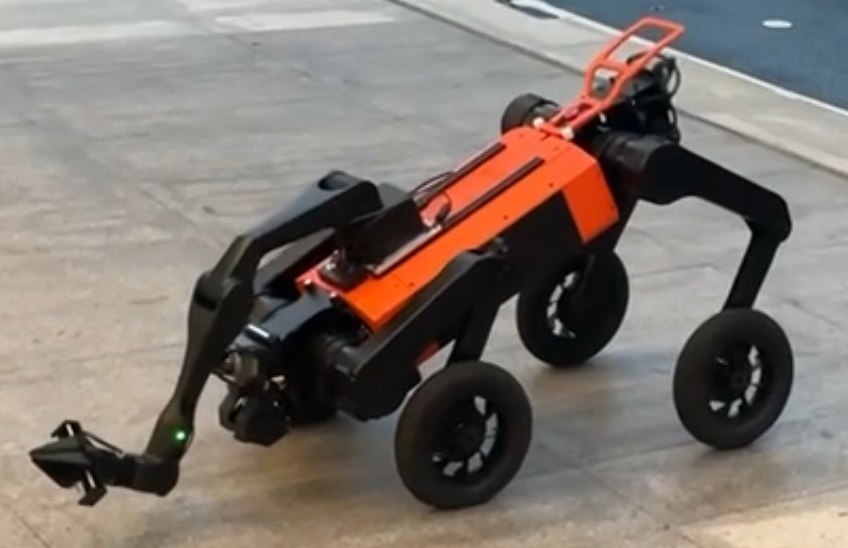}}
    \centerline{(a) manipulation phase}
\end{minipage}%
 \begin{minipage}{0.25\linewidth}
    \centerline{\includegraphics[height=3cm,width=\linewidth]{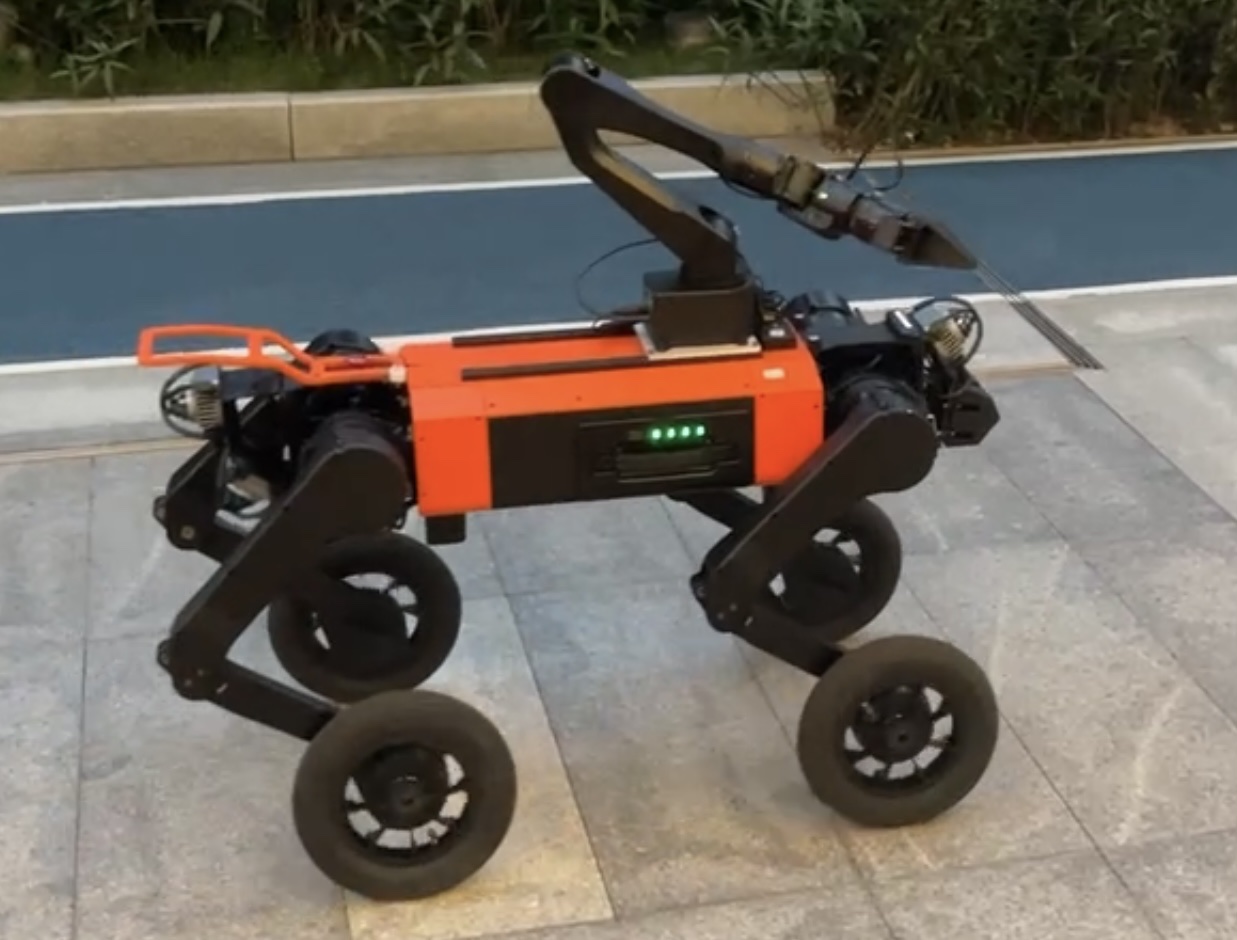}}
    \centerline{(b) locomotion phase}
\end{minipage}%
\begin{minipage}{0.25\linewidth}
     \centerline{\raisebox{0.08cm}{\includegraphics[height=3cm,width=\linewidth]{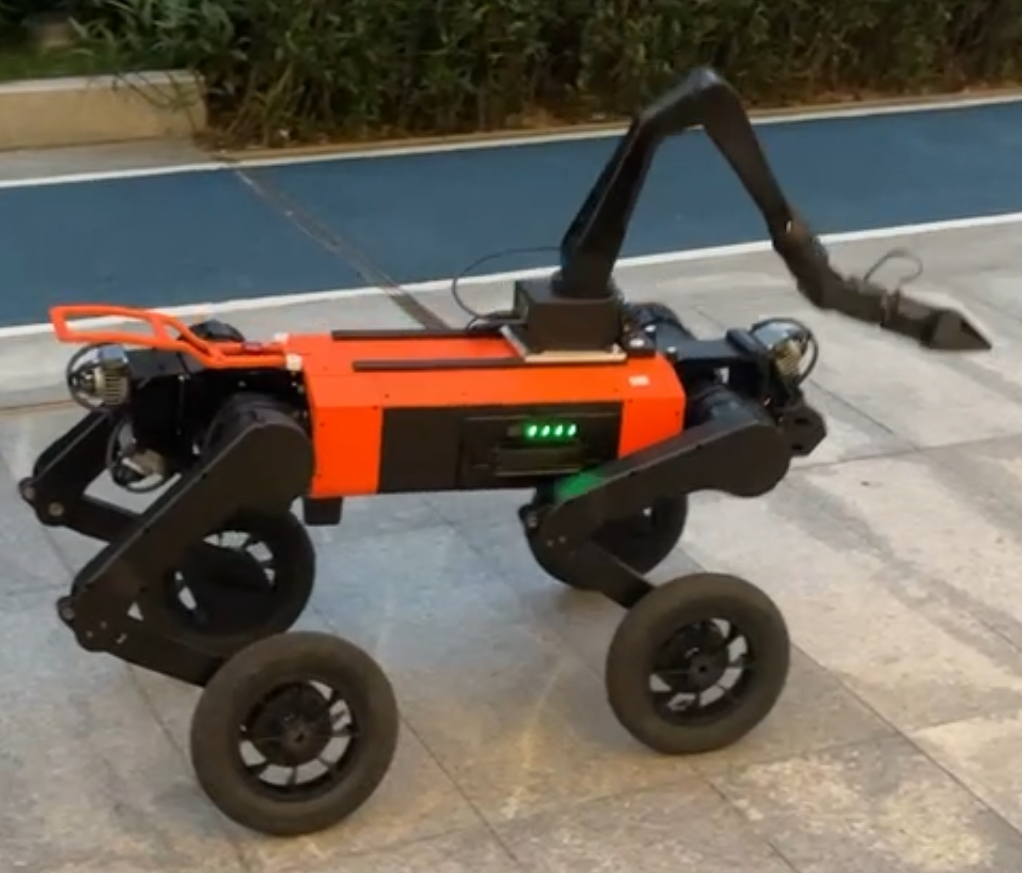}}}
    \centerline{(c) transition of loco. and mani}
\end{minipage}%
\begin{minipage}{0.25\linewidth}
    \centerline{\includegraphics[height=3cm,width=\linewidth]{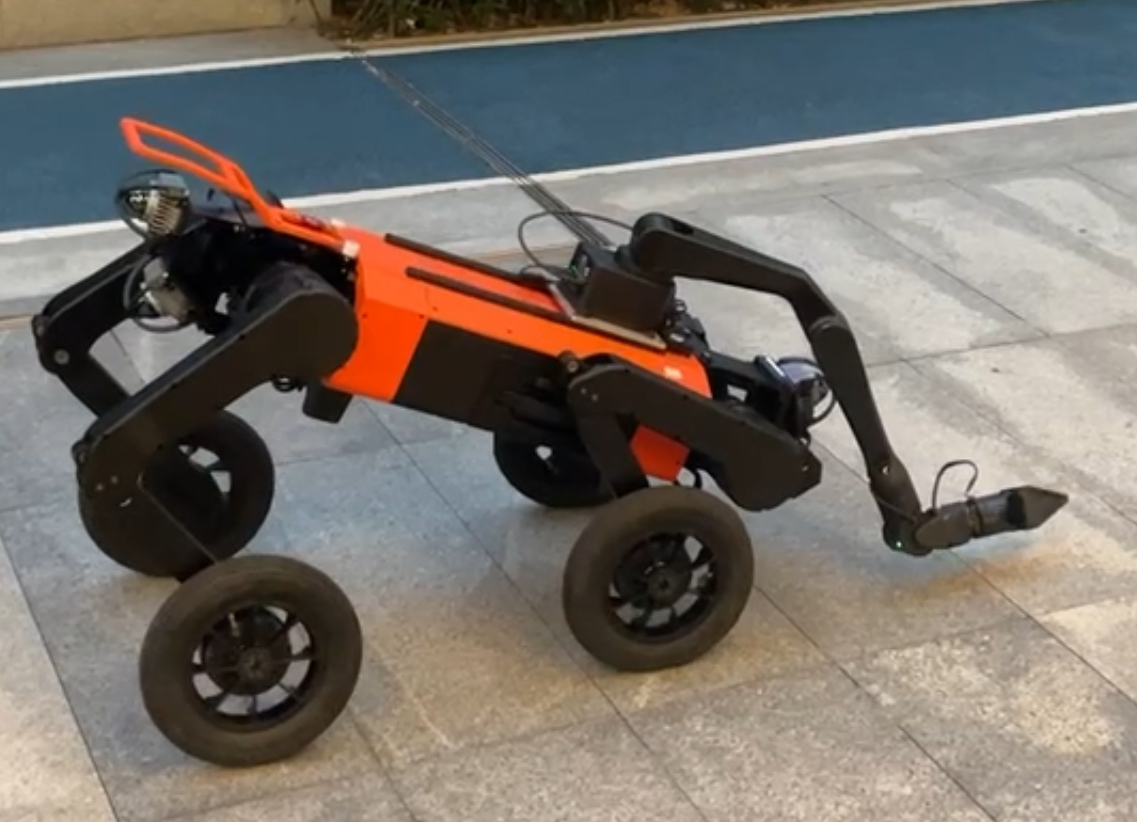}}
    \centerline{(d) manipulation phase}
\end{minipage}
  \caption{Smooth loco-manipulation: (a) At the beginning of the experiment, the EE target is nearby, triggering its manipulation phase. (b) The EE target is suddenly moved far away from the robot. This causes the robot to focus on locomotion to approach the target. (c) As the robot approaching to the target pose, it gradually transit from locomotion to manipulation (d) Once back in the manipulation phase, the robot begins to interact with the target again.}
  \label{fig: mobile manipulation task}
\end{figure*}
\subsubsection{(RFM w/o Reward Prioritization})\label{exp: prioritized reward design}
This ablation study highlights the critical role of RP in our RFM framework for achieving hierarchical relationship among reward terms. RP is essential for imposing regularization during task execution. Without RP, the nominal deviation of robot see an excessive increase, which implies the increase of failure possibility.


\subsubsection{(RFM w/o Enhancement)}\label{exp: Cumulative-based Penalty}
We compare the performance of the agent trained with and without enhancement to assess its impact on precision in both position and orientation tracking. Although the agent trained without enhancement demonstrates lower power consumption, deviation and joint acceleration, they fail to conduct precise and reliable manipulation, which is one of the most crucial aspect for loco-manipulation.

\subsubsection{(Standard PPO w/o RFM)}
This baseline excludes RFM totally and only relies on ``weighted sum''. It is noticeable that there are 8 additional parameters to be tuned compared to RFM. We select the optimal parameters we obtain during a two-day tuning. Nevertheless, w/o RFM still suffers from failure issue, excessive nominal deviation and large joint acceleration. Moreover, the tracking error of w/o RFM is among the worst in Tab~\ref{tab:ablation study}
\subsection{Hardware Validations}
\subsubsection{A whole loco-manipulation task process}
In this experiment, we demonstrate the entire smooth loco-manipulation process, as shown in Fig~\ref{fig: mobile manipulation task}. This cyclical process showcases the integration of locomotion and manipulation. When the target is distant, the agent mainly focus on locomotion task. One can see that WQM keeps the manipulator lying on the base during locomotion to ensure the efficiency of locomotion. When approaching the target, the agent presents smooth transition from locomotion to manipulation, and finally conducts precise manipulation.

\subsubsection{Loco-manipulation tracking error}
We aim to quantify the tracking precision in real world though this experiment. The robot is commanded to track several random individual points and several circular trajectories. We measured the average tracking error as the final result to show the tracking performance. As shown in Table~\ref{tab:experiment tracking performance}, the tracking error is minimal, outperforming the results reported by \cite{Arm2024} and \cite{Fu2023a}. The mathematical explanation of tracking error metrics is the same as those in Ablation Study.
\begin{table}[!htbp]
    \centering
    \caption{Tracking error performance in real world across difference tracking targets.}
    \begin{tabular}{lccc}
        \toprule
        & \multicolumn{1}{c}{EE pos. (m) $\downarrow$} & \multicolumn{1}{c}{EE ori. (rad) $\downarrow$} & \multicolumn{1}{c}{SE(3) $\downarrow$} \\
        \midrule
        fixed points & $0.028\pm0.019$ & $0.089 \pm 0.073$ & $0.145$ \\
        spatial circle & $0.048 \pm 0.022$ & $0.085 \pm 0.025$ & $0.181$ \\
        \bottomrule
    \end{tabular}
    \label{tab:experiment tracking performance}
\end{table}

\subsubsection{Teleoperation to pick up garbage}
In this experiment, we demonstrate how the robot performs a garbage collection task using a simple control interface. Specifically, we use a Quest joystick to send 6-dimension pose commands as the EE target as shown in Fig~\ref{fig:showcase}. Our experiments in garbage collection across both rough and smooth terrains, illustrate our system's omni-directional tracking capabilities and advanced loco-manipulation. Please refer to \href{https://clearlab-sustech.github.io/RFM_loco_mani/}{this website} for more experimental videos.


\section{Discussion and limitation}
We proposed a novel structure named RFM to efficiently fusion all reward terms in a non-linear manner with special functionalities. We demonstrate the critical role of RFM in the loco-manipulation problem under WQM platform. Our agent shows whole-body coordination between wheels, legs, and the arm to fully exploit the capability of the WQM hardware. The trained policy only requires a 6D EE pose command, then the policy will automatically determine the whole-body movement to track such a target. Our proposed policy exhibits state-of-the-art tracking accuracy for the 6D EE tracking task. For future research, it is critical to provide our agent with perceptive information. Since our end-to-end method excludes any manual instruction for base, it obviously requires a more powerful policy to coordinate whole-body motion in some scenarios such as confined space or obstacle-in-way. This can only fulfill if our policy is equipped with perceptive input. Finally, we believe that the methodology proposed in this paper can shine in various tasks and represents an important step towards future development.

\bibliographystyle{IEEEtran}
\bibliography{RAL/RAL2024}  

\end{document}